# MGAN-CRCM: A Novel Multiple Generative Adversarial Network and Coarse-Refinement Based Cognizant Method for Image Inpainting

Nafiz Al Asad[1], Md. Appel Mahmud Pranto[1],
Shbiruzzaman Shiam[1], Musaddeq Mahmud Akand[2],
Mohammad Abu Yousuf[2][†], Khondokar Fida Hasan[3],
Mohammad Ali Moni[4,5]

[1]Department of Information and Communication Technology, Bangladesh University of Professionals, Mirpur Cantonment, Dhaka, 1216, Bangladesh.
[2]Institute of Information Technology, Jahangirnagar University, Savar, Dhaka, 1342, Bangladesh.
[3]School of Professional Studies, Cybersecurity Discipline, University of New South Wales (UNSW), 37 Constitution Avenue, Reid, Canberra, ACT, 2600, Australia.
[4]AI & Digital Health Technology, AI and Cyber Futures Institutes, Charles Sturt University, Bathurst, NSW, 2795, Australia.
[5]AI & Digital Health Technology, Rural Health Research Institute, Charles Sturt University, Orange, NSW, 2800, Australia.

Contributing authors: nafizalasad007@gmail.com;
amhpranto@gmail.com; snshiam9819@gmail.com;
arafakond97@gmail.com; yousuf@juniv.edu; fida.hasan@unsw.edu.au;
m.moni@uq.edu.au;
[†]These authors contributed equally to this work.

**Abstract**

Image inpainting is a widely used technique in computer vision for reconstructing missing or damaged pixels in images. Recent advancements with Generative Adversarial Networks (GANs) have demonstrated superior performance over traditional methods due to their deep learning capabilities and adaptability across diverse image domains. Residual Networks (ResNet) have also gained prominence for their ability to enhance feature representation and compatibility with other architectures. This paper introduces a novel architecture combining GAN and ResNet models to improve image inpainting outcomes. Our framework integrates three components: Transpose Convolution-based GAN for guided and blind inpainting, Fast ResNet-Convolutional Neural Network (FR-CNN) for object removal, and Co-Modulation GAN (Co-Mod GAN) for refinement. The model's performance was evaluated on benchmark datasets, achieving accuracies of 96.59% on Image-Net, 96.70% on Places2, and 96.16% on CelebA. Comparative analyses demonstrate that the proposed architecture outperforms existing methods, highlighting its effectiveness in both qualitative and quantitative evaluations.

# 1 Introduction

Image inpainting is used to restore an image's missing elements without altering the image's original design or textual content. It may be used to modify a picture by eliminating or altering undesired features. Researchers in the fields of computer vision and pattern recognition have taken an avid interest in the process of picture inpainting. It generates realistic data to patch up photos or edit out unwanted elements [1]. Inpainting images is a rapidly developing topic of image processing. One of the primary aims of image restoration is to insert fabricated content into otherwise blank or masked parts of a picture. Many fields are finding practical applications for pictures in painting, such as medical image processing, post-processing in PhotoShop, restoring old books, and more. Hence, it proves the importance of studying the representation of visual elements within the visual arts. A significant problem with picture inpainting is the complexity of real images. There will be apparent fuzzy phenomena in the corrected image and at the boundary between the original and repaired portions. Image inpainting is a technique used in computer vision to restore missing information by reconstructing missing pixels or whole areas based on contextual knowledge. Image manipulation, hidden item recognition, following a target, and intelligent escalation of creativity are all used in image inpainting [2]. Apart from traditional blind inpainting, it has a strong alternative, guided inpainting, which allows users to give explicit guidance or masks while inpainting. Guided inpainting of this sort can guarantee much better control of the results. The generated content would be by the surroundings, making it perfect for tasks like object removal and detailed restoration. In recent years, machine learning and computer vision-based models have been used in many areas such as bone shape reconstruction, image steganography, image restoration, image enhancement, etc. [3–15].

---

Natural images are complicated and diverse; thus, in general, inpainting work needs the development of content pixels and texture patterns and the guarantee of visual authenticity and perceptual plausibility in the outcomes achieved. As a result, images with complicated semantics, high resolution, and significant irregular portions continue to pose a challenge. Even after decades of research, it remains a complex topic in computer vision and graphics. Several international studies have been performed for image inpainting using different computer vision algorithms. Several studies have been conducted to solve and improve image inpainting quality, such as convolutional neural networks, contextual attention or partial convolution [16–22]. Yu et al. [16] recommend a novel contextual layer to pay attention explicitly to relevant features. Even Yan et al. [17] introduce Sift Net, a cutting-edge technology. However, some problems need to be solved in their studies, like not emphasizing high-resolution images, painting complex sparse designs, and the complexity of items such as people and animals to inpaint, artefacts in result, etc. Generative adversarial network (GAN), patchGAN, and spectral normalization techniques are also used by several researchers for image inpainting [23–31]. GAN-based approaches get better results compared to others. Ugur Demir and Gozde Unal. [23] introduce dilated and interpolated convolutions to ResNet, in addition to a complete end-to-end training network aimed at tackling high-resolution picture inpainting. Mohamed Abbas Hedjazia and Yakup Genca. [25] employ gated convolution to train a multiple object selection procedure for every stream at each spatial point across all stages. A pluralistic image completion approach is proposed by Zheng et al. [31] that generates multiple plausible solutions for masked images. Following this, two parallel paths were considered, one for reconstructing the original image and one for generating diverse completions. Here, incorporating GANs, along with short- and long-term attention layers, may improve consistency in appearance from both near and far contexts. However, this method also faces difficulty for highly complex or irregular structures. Sola and Gera et al. [32] proposed the Expression-Conditioned GAN (ECGAN), which leverages both mask segmentation and expression labels to reconstruct expressive masked faces. While effective in generating realistic expression-consistent images, ECGAN can struggle with complex expressions and may require fine-tuning to handle diverse facial structures accurately. Wang et al. [33] proposed a Dual-Path Image Inpainting framework with Auxiliary GAN Inversion, which enhances feed-forward inpainting with semantic priors from GAN inversion. While effective for complex inpainting tasks and large missing areas, this method can face limitations with alignment issues and computational demands for high-resolution images, even with the novel deformable fusion module aimed at improving feature alignment between the dual paths. Yildirim et al. [34] proposed a GAN inversion-based inpainting approach combining encoded erased-image features with random latent codes to produce diverse results. This method achieves varied outputs but needs help with detail fidelity and alignment in large missing areas, even with added gating and skip connections to improve consistency. Many more authors have tried a different kind of GAN-based architecture. Still, a few problems exist to resolve, like locally and globally inconsistent raw image completion, the robustness of highly textured areas, colour distortion for object removal cases, etc. While CNNs have long been the gold standard for image inpainting, transformer-based models have gained significantly on several of their inadequacies. Li et al. [35] proposed the Mask-Aware Transformer (MAT), which is efficient in modeling long-range dependencies but has limited performance about fine textures and, for larger images, is computationally prohibitive. Wan et al. [36] combined transformers and CNNs to improve both structure and detail, but their approach is less practical for larger-sized images due to enhanced complexity. Next, Yu et al. [37] presented BAT-Fill, which uses a bidirectional autoregressive transformer. Unlike unidirectional approaches, BAT-Fill leverages a BERT-like way of modeling contextual information from both directions to enhance inpainting results regarding diversity and fidelity. Ko et al. [38] proposed the CMT, which adapts dynamically to missing information. Still, even this model cannot avoid leaving more significant or irregular gaps with a few annoying side effects. While good results using these models are achieved for complex structures, there are several open problems related to texture preservation and efficiency. In addition to these works, several works exist based on spatial pyramid dilation, recurrent feature reasoning, fast Fourier convolution, etc. [39–42]. Li et al. [39] propose an SPD block for various masks. Again, Li et al. [40] propose working with a Recurrent Feature Reasoning (RFR) algorithm for deeper pixel values. However, their works have a few unresolved problems, such as embedding the hole regions, facing limitations of progressive inpainting, etc. A hybrid method based on several works is accomplished to improve the quality of image inpainting [43–45]. Yang et al. [43] present an optimization method to simulate missing image portions while simulating a local texture restriction. Ran et al. [44] work on structure reconstruction, which is the source of global structure data and generates edge-preserved smooth images. We can identify some unresolved problems in their work, like using irregular masks, vertical fence structures remaining hidden by the objects, etc.

Reviewing previous works in the field of image inpainting, some unresolved problems are found, such as inpaint complex sparse design, artifacts in the final product, locally and globally inconsistent raw images, the embedding of the hole regions, colour distortion, inpaint when irregular masks, etc. To overcome these problems, this paper proposes a model based on multiple generative adversarial networks and coarse-refinement to inpaint both face and natural images and inpaint the place of object removal by resolving existing problems.

The primary objective of this study is to simplify the image inpainting task into two distinct steps: *first*, predicting the overall structure of the missing region, and *second*, creating an image refinement layer based on the predicted structure. The study employs ResNet, an artificial neural network often incorporating nonlinearities (ReLU) and batch normalization. This choice helps avoid the vanishing gradient issue by developing deeper networks and determining an optimal number of layers. The proposed model leverages a multifaceted approach, integrating various methods, models, and architectures. Doing so systematically addresses the challenges identified in prior works, ultimately achieving high-quality inpainting results for various images, including faces, natural scenes, and object-removed regions.

Overall, our main contributions are as follows:

- A novel hybrid architecture is proposed that seamlessly integrates the existing Co-Modulation GAN (Co-Mod GAN) and Fast Resnet- Convolutional Neural Network (FR-CNN) structures with our designed Transpose Convolution-based GAN (TcGAN) framework to address state-of-art challenges in image inpainting.
- To resolve complex sparse design problems for blind inpainting (Blind inpainting refers to a type of image inpainting process where missing or damaged portions of an image are filled in or restored without having access to the ground image), this study proposes a Transpose convolution-based GAN architecture.
- After object removal from the image using only the FR-CNN framework, the resultant image contains artefacts. To remove the artefacts from the resultant image, this paper proposes a refinement layer, which is the Co-ModGAN framework. This framework addresses the local and global inconsistencies in the problem of raw images.
- To resolve the color distortion problem, this study uses the Co-Mod GAN framework, which polishes the resultant image coming from both transpose convolution-based GAN and FR-CNN frameworks. The proposed hybrid framework can excel while inpainting using irregular masks.

The rest of the paper is organized as follows: Section 2 includes an exploration of the datasets we used, such as CelebA, Places2, and Image-Net, and a total explanation of the architecture. Section 3 provides the qualitative results of our approach, the quality of the generated texture, and the scalability. Again, in Section 4, we have shown both qualitative & quantitative comparisons and the efficiency of our approach. And finally, Section 5 includes the conclusion.

## 2 Methodology

In this paper, three models are used together to in-paint and remove an object and then inpaint.

### 2.1 Dataset Exploration

Three datasets are used, CelebA [46], Places2 [47], and Image-Net [48], for the training and validation of the three different models. Among them, the CelebA dataset contains 200K images of size 256x256, Image-Net contains around 1.3M images of size 128x128, and Places2 contains 30K images of size 512x512. All data sets are divided into two groups in an 80-20% ratio for training and validation, respectively.

### 2.2 Proposed Architecture

The proposed hybrid architecture is comprised of three models: Our designed Transpose Convolution based Generative Adversarial Network framework with pre-existing architectures Fast ResNet- Convolutional Neural Network (FR-CNN) and Co-Modulation Generative Adversarial Network (Co-Mod GAN). Our intended work is summarized in Fig. 1, which provides a fully convolutional GAN architecture for image inpainting, which produces locally and globally consistent raw image completion that aims to fill in missing or damaged parts of an image in a way that maintains both local details and overall global coherence.

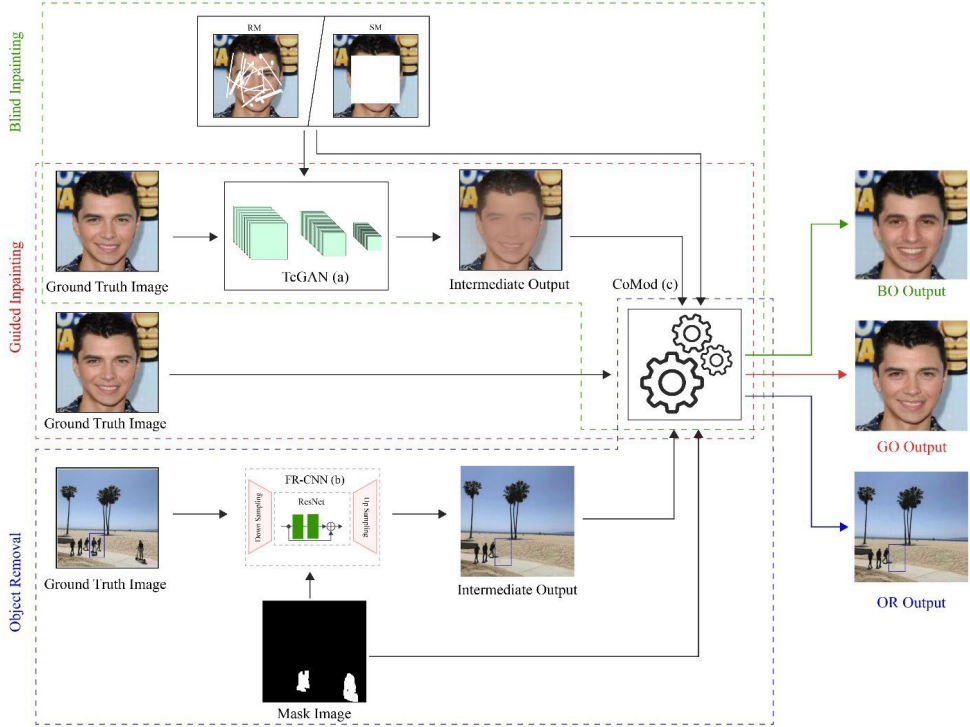

**Fig. 1**: Overview of Proposed Architecture: comprised of three models (1) Transpose Convolution based GAN, (2) FR-CNN & (3) Co-Mod GAN for both blind & guided inpainting and also for object removal.

### 2.2.1 Transpose Convolution Based Generative Adversarial Network (TcGAN)

The Transpose Convolution-Based GAN (TcGAN) is designed as a coarse network to generate the texture and structure of missing regions, particularly in scenarios involving complex sparse designs which is shown in Fig. 2. This network serves as an intermediate stage for both blind and guided inpainting, providing a foundational layer of the inpainting process that subsequent models further refine.

Although transposed convolutions are commonly used in existing GANs, TcGAN introduces significant innovations. Specifically, we incorporate advanced stabilization techniques such as spectral normalization and residual connections to enhance training stability and prevent mode collapse. Additionally, the partial convolutional layer in TcGAN includes a mask update step, ensuring that the convolution operation is both masked and renormalized, thereby focusing on valid data during inpainting. Renormalization adjusts the convolution output based on the number of valid pixels in the convolution window, ensuring consistent results by accounting for the masked areas. This process ensures that the model maintains high performance even when working with irregular or complex missing regions.

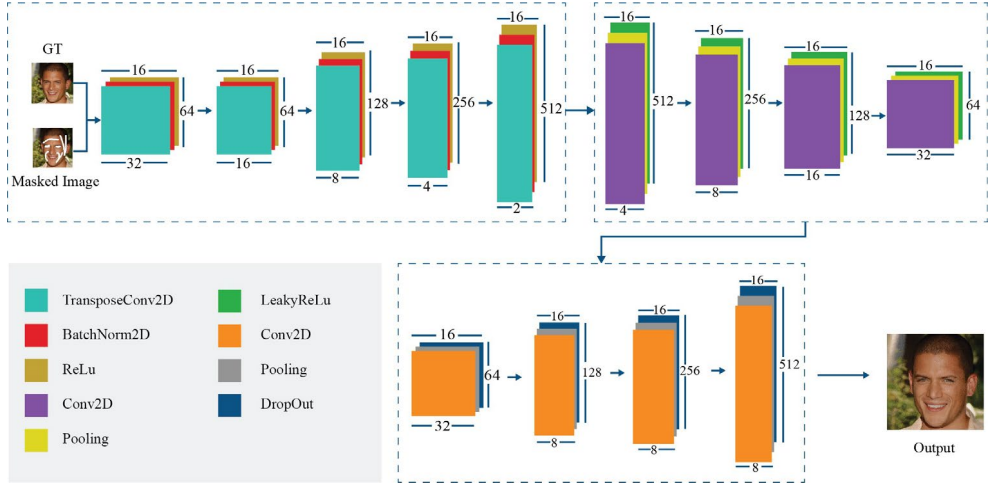

**Fig. 2**: Architecture of Transpose Convolution based GAN: comprised of Generator & Discriminator networks. Ground truth and masked images are the inputs of the generator. Output from the generator is again inserted into the discriminator to get a structure reconstructed image as the final output.

*Generator*

The generator in TcGAN is responsible for producing a preliminary reconstruction of the missing image regions. It comprises down-sampling and up-sampling blocks, which are detailed as follows:

- **Down-Sampling Block:** This block begins by applying 2D convolutions with a 4x4 kernel, a 2x2 stride, and 1x1 padding to encode the input image into a lower-dimensional feature space. The convolution operation is defined as:

$$\text{Out}(N_i, C_{out_j}) = \text{bias}(C_{out_j}) + \sum_k^w \text{weight}(C_{out_j}, k) * \text{input}(N_i, k) \quad (1)$$

- **Batch Normalization:** Following the convolution, a batch normalization process is applied to the network activations to ensure they have a mean of 0 and a variance of 1:

$$y = \frac{x - \mathsf{E}[x]}{\sqrt{\text{Var}[x] + \epsilon}} \times \gamma + \beta \quad (2)$$

Here, $\gamma$ and $\beta$ are learnable parameters, and $\epsilon$ is a small constant added for numerical stability.

- **Leaky ReLU Activation:** The activation function used in the down-sampling block is the Leaky ReLU, which addresses the "dying ReLU" issue by allowing a small, non-zero gradient when the unit is not active:

$$\text{LeakyReLU}(x) = \max(0, x) + \text{negative\_slope} \times \min(0, x) \quad (3)$$

- **Up-Sampling Block:** After encoding, the up-sampling block utilizes 2D transposed convolutions to reconstruct the image from the encoded features. The dimensions of the transposed convolution output are computed as follows:

$$H_{out} = (H_{in} - 1) * stride[0] - 2 * padding[0] + dilation[0] * (kernel\_size[0] - 1) + output\_padding[0] + 1 \quad (4)$$

$$W_{out} = (W_{in} - 1) * stride[1] - 2 * padding[1] + dilation[1] * (kernel\_size[1] - 1) + output\_padding[1] + 1 \quad (5)$$

Dilation in Equations 4 and 5 refers to the spacing between kernel elements, which controls the receptive field of the convolution operation, allowing the network to capture a broader range of features during the up-sampling process.

*Discriminator*

The discriminator in TcGAN is tasked with distinguishing between authentic images and those generated by the generator. It evaluates both local and global consistency to ensure that the inpainted regions are indistinguishable from the surrounding accurate image content.

- **Input Handling:** The discriminator receives the generated image from the generator alongside the ground truth image. Both images are processed through a series of convolutional layers that progressively refine the network's ability to differentiate between real and generated content.
- **Training Process:** In TcGAN, the generator and discriminator are trained simultaneously in an adversarial framework. The generator aims to produce realistic images that fool the discriminator, while the discriminator learns to classify the authenticity of the images it receives accurately. This adversarial training continues iteratively, with the generator and discriminator updating their gradients to minimize their respective loss functions. In addition, we employ an Upgraded Gradient Technique. This technique integrates spectral normalization, a stabilization method proposed by Miyato et al. [49], which is used to stabilize gradient flow during training, ensuring that both networks maintain consistent performance and avoid problems such as gradient vanishing or explosion. This is done through backpropagation, allowing the networks to improve with each training iteration.

**Implementation Details and Network Parameters:** We carefully chose specific network parameters to optimize performance during training. The learning rate for the generator and discriminator is set to 0.0002, which balances training speed and stability well. We used a batch size of 64 to ensure the model can generalize well across different input patterns. The latent vector dimension is set to 128, providing enough representation without risking overfitting. The model was trained over 100 epochs to give it plenty of time to learn more intricate patterns in the data. These choices were based on extensive testing and tuning for the best results.

Algorithm 1 provides a detailed breakdown of the training process for TcGAN, demonstrating how the generator and discriminator gradients are updated during each iteration:

This detailed algorithm ensures that TcGAN is effectively trained to generate high-quality coarse images that are passed on for further refinement.

### 2.2.2 Fast ResNet-Convolutional Neural Network (FR-CNN)

The Fast ResNet-Convolutional Neural Network (FR-CNN) framework adds or removes objects within an image, effectively filling in missing regions by applying deep feature-level guidance for the encoder layers through a distillation-based strategy. The output of this model is then fed into the Co-Mod GAN for further refinement, ensuring that the final image is consistent and visually accurate.

FR-CNN operates similarly to the Residual Convolutional Neural Network (R-CNN) architecture, with specific enhancements tailored for inpainting tasks. The architecture of the FR-CNN model is shown in Fig. 3, showing how ground-truth and masked images are processed through the ResNet module, involving both down-sampling and up-sampling operations. The intermediate output from the ResNet module is then passed through the FR-CNN block to produce the final output image, which the Co-Mod GAN then refines.

The FR-CNN model processes the ground truth (GT) image along with a user-guided masked image to construct a convolutional feature map. The following steps outline the key operations within the FR-CNN framework:

- **Convolutional Feature Map Construction:** The GT image and the masked image are combined, and a feature map is generated through a series of convolutional layers. This map captures the essential features of the image, both from the existing content and the masked regions.
- **RoI Pooling and Fully Connected Layers:** Regions of interest (RoI) are extracted from the convolutional feature map. These regions are warped into squares and reorganized into a fixed size using a RoI pooling layer. This output is then fed into fully connected layers to further process and refine the feature representations.
- **Softmax Layer for Region Classification:** Based on the RoI feature vector, a softmax layer is used to predict the class of the proposed region and to calculate the offset values for the bounding box, which defines the spatial extent of the region being modified.
- **Mask Convolution:** In our model, the convolution of the mask into the RGB picture is done using a 4-channel tensor. This approach ensures that the inpainting

---
**Algorithm 1** Algorithm for Transposed Convolution based GAN
---
**Input: Ground truth and mask images**
**Output: Regenerated images with texture and structures**
 1: Load Ground truth images and corresponding Mask images
 2: Initialize G (generator) and D (discriminator) networks
 3: Set training parameters: number of training iterations (epochs), batch size (k)
 4: *LOOP Process*
 5: **for** epoch in range (number of training iterations) **do**
 6:     **for** step in range(k) **do**
 7:         noise_samples = sample_m_noise_samples z(1)..........................z(m)
 8:         example_images = sample_m_example_images x(1). ..............x(m)
 9:     **end for**
10:     **Updating the discriminator's stochastic gradient:**
11:     **for** i in range(m) **do**
12:         real_image = example_images[i]
13:         fake_image = generator(noise_samples[i])
14:         discriminator_real_score = D(real_image)
15:         discriminator_fake_score = D(fake_image)
16:     **end for**
17: **end for**
18: **Compute discriminator loss and update gradients:**
19: discriminator_loss = compute_discriminator loss(discriminator real_score, fake_score)
20: update_discriminator_gradients(discriminator_loss)
21: **Sample another minibatch of noise samples for generator:**
22: noise_samples = sample_m_noise_samples z(1)................. z(m)
23: **Update generator's stochastic gradient:**
24: **for** i in range(m) **do**
25:     fake_image = generator(noise_samples[i])
26:     discriminator_fake_score = D(fake_image)
27: **end for**
28: **Compute generator loss and update gradients:**
29: generator_loss = compute_generator loss(discriminator_fake_score)
30: update_generator_gradients(generator_loss)
31: **Transposed Convolution to upscale generated content**
32: upscaled_fake_image = transposed_convolution(fake_image)
---

models have access to the entire image as soon as possible, thereby preventing unnecessary build-up of context layer by layer. This reduces computational overhead and optimizes the use of model parameters. The mask used in our model is created based on the regions of the image that require inpainting. For user-guided inpainting, the user manually selects the region, which is then converted into a binary mask, with the inpainting area marked as 1 (white) and the rest as 0 (black). For blind inpainting, a pre-defined or randomly generated mask selects regions within the image, aligning well with the areas needing reconstruction.
- **Fourier Transformation and Convolution:** The FR-CNN integrates a local branch that utilizes standard convolutions and a global unit that performs a channel-wise real Fourier Transformation. This global unit transforms the input into the frequency domain, where a 3x3 convolution is applied to recover the spatial structure of the input data. The outputs from both branches are then combined channel by channel.

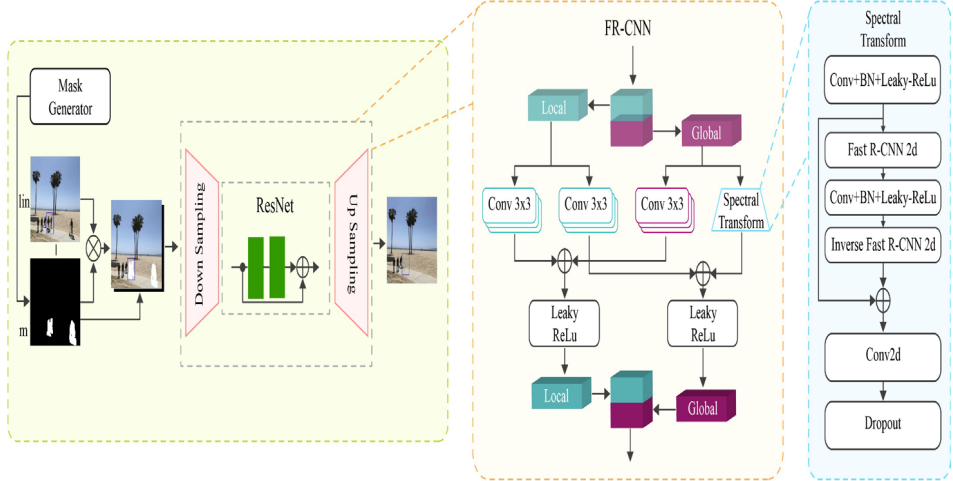

**Fig. 3**: Architecture of Fast ResNet-Convolutional Neural Network (FR-CNN): Ground truth and masked image are combined to serve as the input for the ResNet module through down-sampling and up-sampling operations. The intermediate output from the ResNet module is passed through the FR-CNN block to produce the final output, which is used for object removal purposes.

**Advantages of FR-CNN:** The FR-CNN model provides the inpainting process with immediate access to the image's global context while capturing periodic features commonly found in images (e.g., human figures, textures like braids, etc.). By combining the production lengths of both local and global processing units, the FR-CNN ensures that the inpainted regions blend seamlessly with the rest of the image, both in terms of content and texture.

The output of the FR-CNN is then fed into the Co-Mod GAN for further refinement, ensuring that the final output is not only structurally accurate but also visually appealing and contextually consistent.

### 2.2.3 Co-Modulation Generative Adversarial Network (Co-Mod GAN)

The Co-Modulation Generative Adversarial Network (Co-Mod GAN) serves as a refinement layer within the network, bridging the gap between conditional and unconditional GANs. This model is designed to enhance the image completion process by combining the precision of conditional modulation with the stochastic generative power of unconditional modulation. By leveraging these capabilities, Co-Mod GAN effectively generates the color and context of images using fragments of visual information, enabling both blind and user-guided inpainting, regardless of whether the Ground Truth (GT) image is provided.

The architecture of Co-Mod GAN is illustrated in Fig. 4.

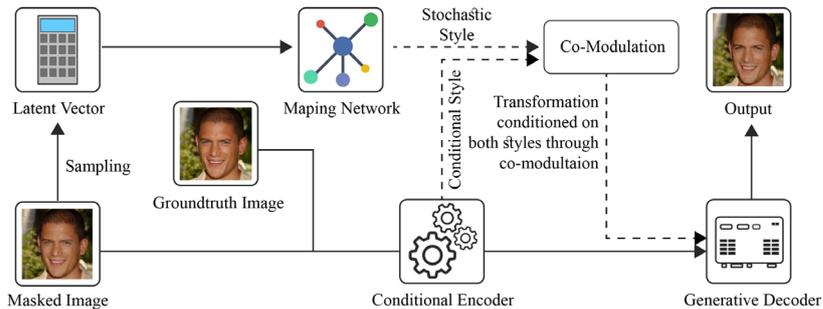

**Fig. 4**: Architecture of CoMod GAN (Refinement Layer): Ground truth and structure-reconstructed image (masked image) from both the Transpose Convolution based GAN and FR-CNN model are sent to the Latent Vector and Conditional Encoder, which ends up in the Generative Decoder after being passed through the Co-Modulation layer.

*Conditional and Unconditional GAN Integration*

Conditional GANs cannot traditionally generate a wide range of outcomes due to their reliance on conditional inputs, which limit the stochastic diversity of the generated images. This limitation becomes increasingly problematic when dealing with large-scale image completion tasks, where the model needs to generalize from limited conditional information.

Co-Mod GAN addresses this issue by incorporating both conditional and unconditional modulation techniques. This dual-modulation approach allows the model to generate images that match the expected output based on the conditional input and retain a high degree of variability and realism. The model employs identical affine transformations on both fashion representations, assuming a linear relationship between the conditional and unconditional factors in the aesthetic space, resulting in immediately striking outputs.

*Linear Correlation and Stochasticity*

In Co-Mod GAN, the assumption of a linear association between the style representations allows for practical trade-offs between image quality and intra-conditioning variety. This trade-off is crucial for achieving high-quality results in large-scale missing patches while maintaining the stochastic generating potential of the model. Unlike traditional GANs that may suffer from external losses when balancing quality and variety, Co-Mod GAN's co-modulation approach preserves both aspects effectively.

The training of Co-Mod GAN utilizes standard discriminator losses rather than the L1 term, fully leveraging its stochastic generating capabilities. This approach ensures that the model can generate high-quality, contextually consistent images even in challenging inpainting scenarios.

*Working Procedure*

The operational process of Co-Mod GAN is straightforward:

- **For User-Guided Inpainting:** We use the original image as the GT and the output of the Transpose Convolution-based GAN as the mask, which is first converted to a grayscale image.
- **For Blind Inpainting:** We use the output of the Transpose Convolution-based GAN as the GT and the same black-and-white mask used by the Transpose Convolution-based GAN is applied.

The input, whether a combination of GT and masked images or just the masked image, is first converted into a latent vector, fed into the mapping network. Simultaneously, the input image is processed directly through a conditional down-sampler. The outputs from both networks are then passed into the GAN up-sampler, where the final refined image is generated.

Co-Mod GAN's integration of conditional and unconditional modulation, coupled with its linear correlation assumption, significantly enhances its ability to perform high-quality inpainting, making it a powerful component of the overall image inpainting framework.

## 2.3 Ablation Studies and Parameter Discussion

In this section, we'll discuss the tests we did to see how important each part of our model is. Our model includes several components: the Transpose Convolution-based GAN (TcGAN), Fast ResNet-Convolutional Neural Network (FR-CNN), and Co-Modulation GAN (Co-Mod GAN). Taking out or changing these components, we can understand how much each contributes to the model's overall performance.

### 2.3.1 Ablation Study Design

We started with a basic version of our model that didn't include any of the unique features we developed. This basic version is just a simple GAN without the transpose convolutions, residual connections, or the refinement layer that we added later.

### 2.3.2 Component Removal/Modification

- **Without TcGAN:** The first step in our ablation study was to remove the Transpose Convolution-based GAN (TcGAN) from the model. TcGAN is designed to handle complex sparse patterns that need inpainting, especially when generating realistic textures and keeping the structure intact. Taking TcGAN out lets us see how much it contributes to the model's ability to rebuild detailed textures and deal with sparsely structured missing areas. The drop in performance in this scenario highlights just how critical TcGAN is for handling these specific challenges.
- **Without FR-CNN:** Next, we tested the model without the Fast ResNet-Convolutional Neural Network (FR-CNN). This component is crucial for object removal and hole filling during the inpainting process. FR-CNN provides deep feature guidance, which is critical for accurately removing objects from images and ensuring the filled regions look consistent. Without this part, the model struggled with maintaining image quality and avoiding the creation of artifacts. This showed us how vital FR-CNN is in keeping the overall image coherent, ensuring that the inpainted regions blend smoothly with the surrounding content.
- **Without Co-Mod GAN:** Finally, we looked at what happens when the Co-Mod GAN is removed. This layer acts as a refinement tool, improving the final image by focusing on color correction, texture generation, and overall appearance. Without Co-Mod GAN, we observed how the model's ability to produce high-quality, visually consistent images was affected. The results showed that without this refinement layer, the images produced lacked the polish and consistency that Co-Mod GAN usually provides, making it clear that this component is essential for maintaining high visual fidelity.

### 2.3.3 Varying Parameters

- **Residual Networks Parameters:** The next set of experiments focused on tweaking critical parameters within the residual networks, which are crucial to the model's overall performance. We experimented with different settings, like the number of layers, the learning rate, and the use of batch normalization. The goal was to find the best balance between performance and computational efficiency. For example, adding more layers to the residual network could improve feature extraction and lead to overfitting if not handled carefully. These experiments allowed us to explore the trade-offs and determine the best parameter settings for our model.
- **Stride and Kernel Size in TcGAN:** Another critical area we explored was how changes in stride and kernel size within the Transpose Convolution layers would affect the model. These parameters control the scale and resolution of the textures generated during inpainting. By adjusting them, we could see how they influenced the quality of the inpainting results, especially when it comes to keeping textures and structures looking natural across different scales of missing regions. For instance, more considerable strides might speed up processing but could result in a loss of finer details, while smaller strides might enhance detail but require more computational resources. These tests were crucial in fine-tuning TcGAN for optimal performance.

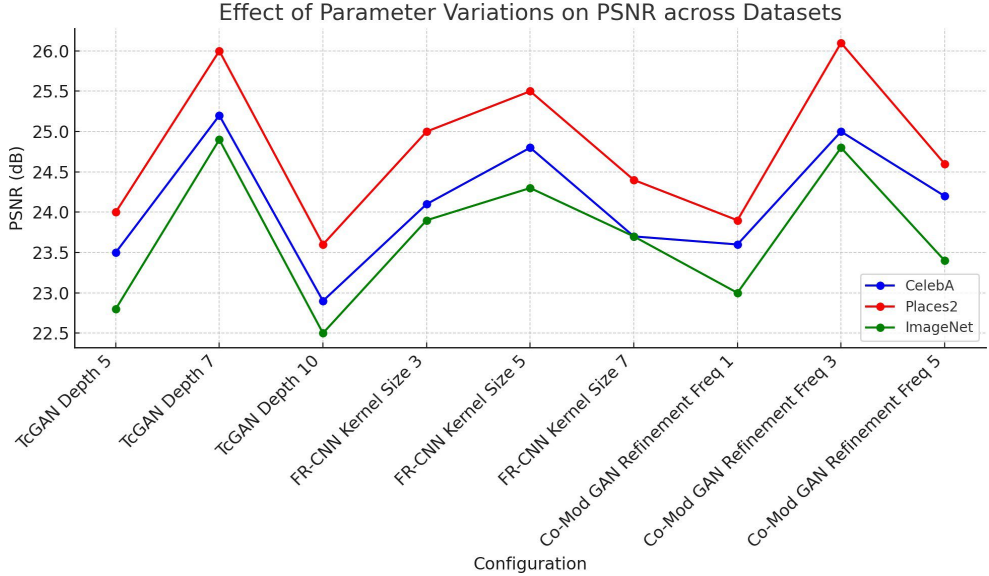

**Fig. 5**: PSNR results for varying TcGAN depths, FR-CNN kernel sizes, and Co-Mod GAN refinement frequencies across CelebA, Places2, and ImageNet datasets. Moderate depth and refinement frequencies yield optimal PSNR.

- **Refinement Frequency in Co-Mod GAN:** Lastly, we discussed how often the Co-Mod GAN is applied within the network. Co-Mod GAN is used to refine the output images, and by changing the frequency of this refinement process, we can see how it affects the smoothness and realism of the final images. Using it more frequently might improve refinement at the cost of processing time, while less frequent use could speed up the process but result in less polished outputs. This analysis helped us determine the most effective way to use Co-Mod GAN within the model.

The impact of these parameter variations on PSNR, SSIM, and MAE across CelebA[46], Places2[47], and ImageNet[48] datasets is illustrated in Fig. 5, 6, and 7. As shown in Fig. 5, PSNR values increase with TcGAN depth, reaching a peak at a depth of 7, particularly on Places2. However, increasing depth further to 10 results in decreased PSNR, especially on CelebA and ImageNet, suggesting that deeper networks may lead to overfitting.

Fig. 6 shows SSIM, with CelebA and Places2 reaching optimal structural similarity at a refinement frequency of 3. However, increasing frequency to 5 reduces SSIM across datasets, indicating that excessive refinement may disrupt texture consistency. Notably, Places2 maintains higher SSIM, likely due to its structured content.

In Fig. 7, MAE trends inversely with PSNR and SSIM. An increased refinement frequency of 5 results in higher MAE values, which suggests excessive refinement may introduce more pixel-level errors. These results emphasize the importance of balanced tuning to maximize inpainting quality across diverse datasets.

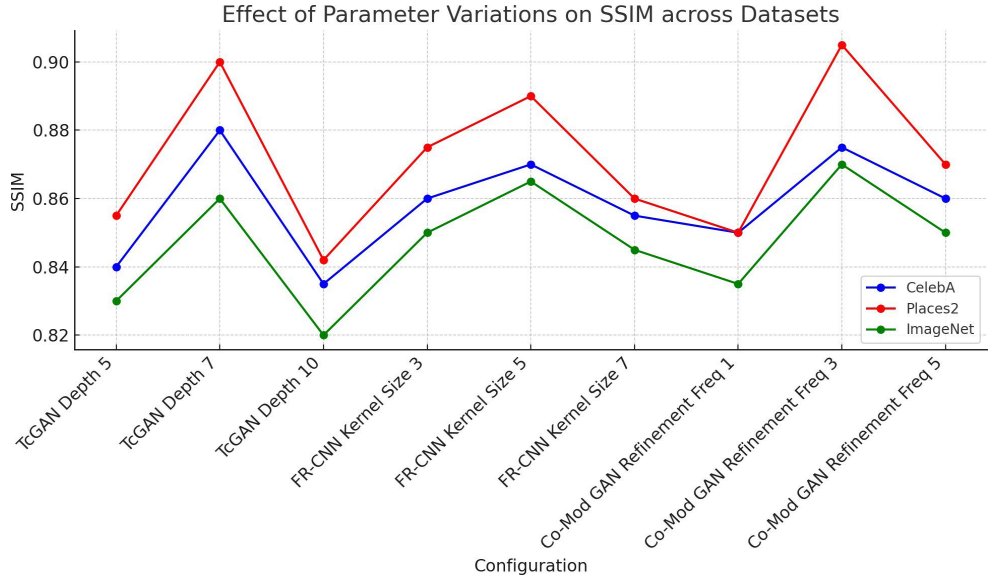

**Fig. 6**: SSIM values across CelebA, Places2, and ImageNet datasets, showing how refinement frequency affects structural similarity. The highest SSIM is achieved at a frequency of 3, with declines at more extreme settings.

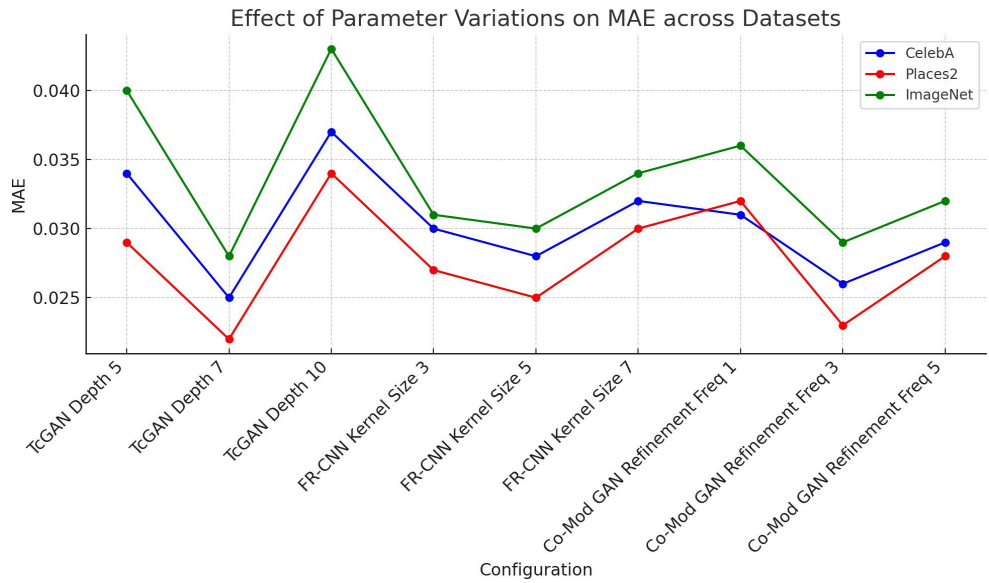

**Fig. 7**: MAE values across datasets for each parameter configuration, indicating an increase in pixel error with higher refinement frequencies.

### 2.3.4 Results and Discussion

We conducted a comprehensive evaluation of each model configuration using standard metrics such as Peak Signal-to-noise ratio (PSNR), Structural Similarity Index Measure (SSIM), and Mean Absolute Error (MAE). These evaluations were performed across the CelebA, Places2, and ImageNet datasets. The results are summarized in Table 1, which provides a detailed comparison between the baseline model, the fully enhanced model, and models with individual components removed.

Table 1: Evaluation Results Across Different Datasets and Model Configurations

| Configuration | CelebA[46] | | | Places2[47] | | | ImageNet[48] | | |
|---|---|---|---|---|---|---|---|---|---|
| | PSNR↑ | SSIM↑ | MAE↓ | PSNR↑ | SSIM↑ | MAE↓ | PSNR↑ | SSIM↑ | MAE↓ |
| **Baseline Model** | 23.45 | 0.842 | 0.034 | 24.02 | 0.856 | 0.029 | 22.89 | 0.832 | 0.042 |
| **w/o TcGAN** | 21.38 | 0.814 | 0.045 | 22.14 | 0.832 | 0.037 | 21.12 | 0.808 | 0.050 |
| **w/o FR-CNN** | 22.05 | 0.827 | 0.039 | 23.05 | 0.844 | 0.034 | 21.99 | 0.819 | 0.047 |
| **w/o Co-Mod GAN** | 22.72 | 0.835 | 0.037 | 23.68 | 0.851 | 0.031 | 22.57 | 0.825 | 0.044 |
| **Full Model** | 25.19 | 0.890 | 0.028 | 26.31 | 0.906 | 0.022 | 24.85 | 0.878 | 0.031 |

- **Effectiveness of TcGAN** Removing the Transpose Convolution-based GAN (TcGAN) led to a noticeable decline in model performance across all datasets. For instance, on the CelebA dataset, the PSNR dropped from 25.19 dB in the full model to 21.38 dB without TcGAN, while SSIM decreased from 0.890 to 0.814. Similar trends were observed on the Places2 and ImageNet datasets, where the PSNR dropped to 22.14 dB and 21.12 dB, respectively, and SSIM values declined to 0.832 and 0.808. Furthermore, the MAE increased significantly, indicating a more significant error in the reconstructed images. Specifically, the MAE rose from 0.028 to 0.045 on CelebA, highlighting the model's struggle to fill in complex patterns without TcGAN accurately. These results demonstrate that TcGAN is crucial for generating realistic textures and maintaining structural integrity in inpainting tasks. The consistent drop in performance across multiple datasets confirms the indispensable role of TcGAN in the model's architecture.
- **Impact of FR-CNN** The exclusion of the Fast ResNet-Convolutional Neural Network (FR-CNN) also significantly impaired the model's performance, particularly in handling object removal and hole filling. The PSNR decreased to 22.05 dB on CelebA, 23.05 dB on Places2, and 21.99 dB on ImageNet, while the SSIM values dropped to 0.827, 0.844, and 0.819, respectively. The MAE values also worsened, increasing to 0.039 on CelebA, 0.034 on Places2, and 0.047 on ImageNet. The resulting images displayed visible artifacts and inconsistencies, emphasizing the essential role of FR-CNN in ensuring smooth and natural-looking inpainting. This part of the study highlights how FR-CNN contributes to maintaining overall image quality by preventing the generation of artifacts, which is critical for the success of the inpainting process.

- **Role of Co-Mod GAN** When the Co-Modulation GAN (Co-Mod GAN) was removed from the model, the results showed a decline in both color balance and texture quality. The PSNR for the CelebA dataset dropped to 22.72 dB, and SSIM decreased to 0.835, indicating a less refined output than the full model. Similar performance drops were observed on the Places2 and ImageNet datasets, with PSNR values of 23.68 dB and 22.57 dB and SSIM values of 0.851 and 0.825, respectively. The MAE also increased, particularly on the ImageNet dataset, where it rose to 0.044. Without Co-Mod GAN, the images lacked the visual polish that this component typically provides, underscoring its importance in the final image processing stage. The study shows that Co-Mod GAN is vital for refining the output to ensure it is structurally sound and visually appealing.

**Parameter Variation Insights:** Through parameter variation experiments, it was observed that specific configurations, such as increasing the depth of the residual network and fine-tuning the stride in TcGAN, led to significant improvements in image quality. These adjustments resulted in higher PSNR and SSIM values, as well as lower MAE across all datasets, validating the design choices made in our model. For instance, the whole model achieved a PSNR of 26.31 dB on the Places2 dataset with an SSIM of 0.906 and an MAE of 0.022, demonstrating the effectiveness of these optimizations.

The ablation studies and parameter variation experiments clearly confirm that each component of our model plays a critical role in delivering high-quality results.

The thorough analysis presented here demonstrates that the model's success depends on each component's careful integration and optimization, with TcGAN, FR-CNN, and Co-Mod GAN contributing significantly to the overall performance.

## 3 RESULTS

In this part, we've tested the suggested multi-gan image inpainting strategy using CelebA, Places2, and ImageNet datasets. Our datasets include 0.5M photos. The approach is tested on an Intel Core-i5 7200U CPU running at 2.5 to 2.81 GHz and an Nvidia GeForce 940mx graphics card. It takes 30 ms on the GPU and 40 ms on the CPU to test.

### 3.1 QUALITATIVE RESULTS

Our proposed system works on guided and blind image inpainting. For inpainting purposes, random masks, center masks, and user-guided masks are all exerted.

Fig. 8a shows the result of guided image inpainting using a random mask. The first column is the Ground Truth (GT), the second column represents random masks, the third column shows the structure reconstruction (SR), and the fourth column shows the resultant output of the proposed work.

Fig. 8b shows the guided image inpainting using a rectangle mask; the first column shows the ground truth (GT) image, the second column shows the rectangular mask, the third one shows structure reconstruction (SR), and the fourth column shows the generated output image.

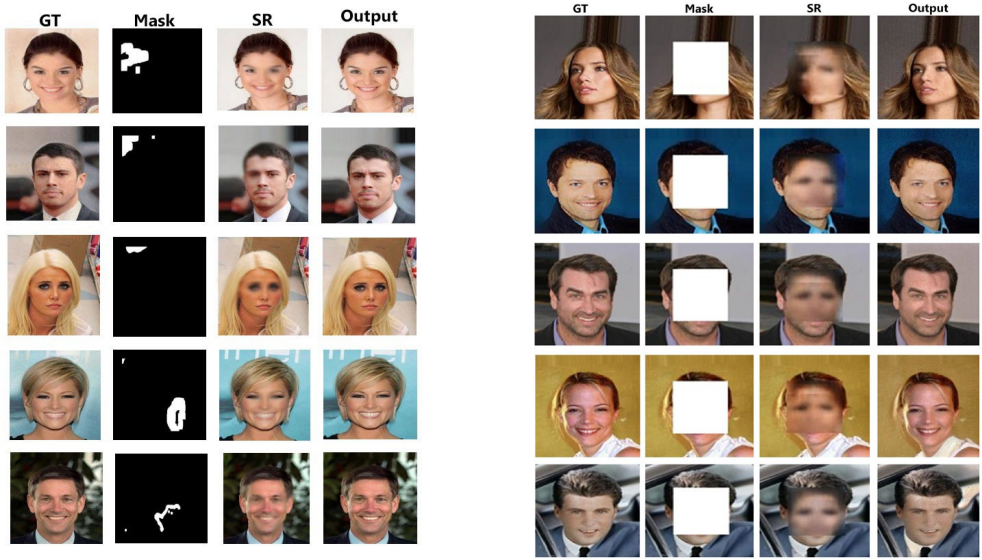

(a) Guided image inpainting using random masks: From left Ground Truth(GT), Mask, Structure Reconstruction (SR), and Output

(b) Guided Image Inpainting for center rectangular masks: From left Ground Truth(GT), Mask, Structure Reconstruction (SR), and Output

**Fig. 8**: Guided image inpainting using random masks and center rectangular maks

This Fig. 9a represents the blind image inpainting using rectangular and user-guided masks. The first column represents the masked image, the second column is the generated output, and the third one is the ground-truth image.

In Fig. 9b, we see the full results of the object removal process. The first column is the "ground truth" picture, the second column is the "masked" image, and the third column is the "produced" output, which reveals that no item was created in the "masked" region.

### 3.2 Analysis of the Generated Texture Quality

To measure the performance of our method, we have calculated the boundary performance accuracy evaluation metrics on the masked edges for the CelebA, Places2, and ImageNet datasets since the boundaries are the heart of the image structure for further evaluation. We have used Canny [50], as it is one of the popular boundary detectors to measure the difference between the boundaries in the reconstructed image and the ground truth image. As seen in Tables 2, we have used multiple evaluation metrics for the reconstructed boundaries of the masked region to analyze the number of reconstructed boundaries in transpose convolution-based GAN. Table 2 shows that our method has recovered the maximum details as it achieved remarkable scores in all evolution metrics. We have also tested our model FR-CNN and Co-Mod network on three datasets to evaluate its performance. Table 3 and Table 3 show that our

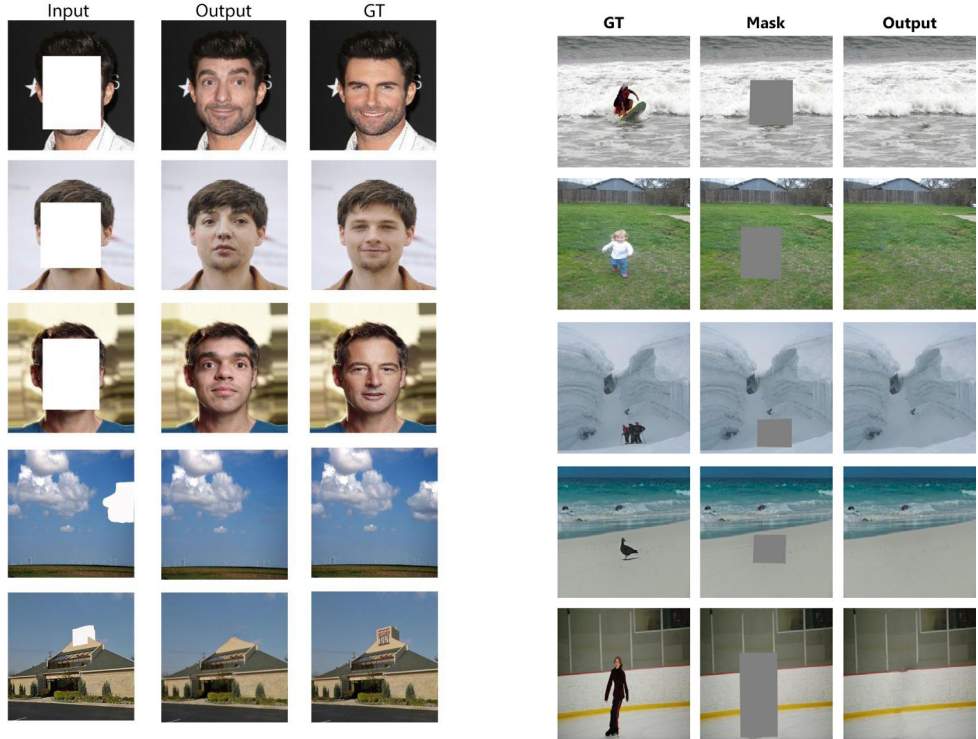

(a) Blind Image Inpainting for center rectangular and user-guided masks: From left Input Image, Model Output, and Ground Truth (GT)

(b) Object Removal and Inpainting using user guided masks: From left Ground Truth (GT), Mask, and Model Output

**Fig. 9**: Blind image inpainting and Object removal & inpainting

**Table 2**: Performance evaluation metrics for transpose convolution-based GAN on CelebA, Places2 and ImageNet datasets

| Datasets | Accuracy (%) | Precision (%) | Recall (%) | F1 Score (%) |
|---|---|---|---|---|
| CelebA | 95.61 | 87.43 | 86.21 | 86.40 |
| Places2 | 96.32 | 86.20 | 85.39 | 87.50 |
| ImageNet | 96.59 | 85.72 | 85.01 | 85.63 |

model scored notably high in all measured performance evaluation metrics. From Fig. 10a, we can find average scores from three different datasets (CelebA, Places2 and ImageNet) for each model performance evaluation metric. From Fig. 10b, we can find the Performance Assessment Measurements on the three different models (transpose convolution-based GAN, Co-Mod GAN, and FR-CNN) average scores of each data set.

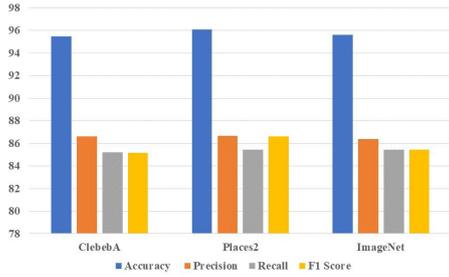
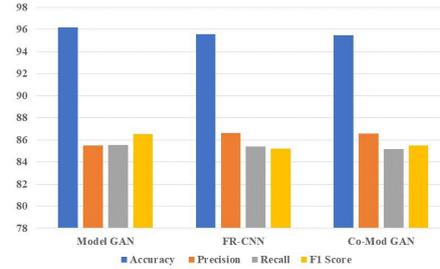

(a) Visual scene Of Performance Evaluation Metrics on average scores of three individual datasets (Celeba, Places2 And ImageNet) for each model.

(b) Visual Scene Of Performance Evaluation Metrics on average scores of the three individual models(Transpose COnvolution based GAN, Co-Mod GAN, and FR-CNN) On each dataset

**Fig. 10**: Performance Evaluation Metrics for each dataset and each sub-model

**Table 3**: Performance evaluation metrics for Fast R-CNN on CelebA, Places2 and ImageNet datasets

| Datasets | Accuracy (%) | Precision (%) | Recall (%) | F1 Score (%) |
|---|---|---|---|---|
| CelebA | 94.61 | 86.10 | 85.12 | 85.02 |
| Places2 | 96.70 | 86.63 | 84.93 | 86.27 |
| ImageNet | 95.36 | 87.16 | 86.10 | 84.36 |

**Table 4**: Performance evaluation metrics for Co-Mod GAN on CelebA, Places2 and ImageNet datasets

| Datasets | Accuracy (%) | Precision (%) | Recall (%) | F1 Score (%) |
|---|---|---|---|---|
| CelebA | 96.16 | 86.34 | 84.23 | 84.01 |
| Places2 | 95.23 | 87.13 | 86.06 | 86.06 |
| ImageNet | 94.95 | 86.27 | 85.25 | 86.36 |

## 3.3 Evaluation Metrices comparisons

To ensure our strategy is scalable, we've performed the three trials shown in Table 5. We've trained and tested the suggested method with a free-form mask to ensure that it can be generalized to new data. Our method does exceptionally well in creating realistic photos with the correct texture despite having the model on invisible people's faces and locations. In the second experiment, we tested random user-guided masks on Places2, ImageNet, and CelebA datasets. We have taken the mask to different locations. We've used the random mask and the rectangular mask. Although it is a more challenging task than blind inpainting, our method has generated notably visually plausible results both in Places2, ImageNet, and CelebA datasets. It has reconstructed the image with proper texture, both uniform and non-uniform backgrounds. In the third experiment, we measured the scalability of the object removal approach on three datasets. As seen in Table 5, our approach has done noteworthy scores in the three metrics. Our approach can remove multiple unwanted objects from any location in the image and inpaint that region with appropriate texture and color.

**Table 5**: Demonstration of parameters: MAE ↓, SSIM ↑ & PSNR ↑ with datasets (Places2, ImageNet, CelebA) for blind inpainting, guided inpainting & object removal

| Task | Metric | Places2 | ImageNet | CelebA |
|---|---|---|---|---|
| Blind Inpainting (User Guided Masks) | MAE ↓ | 0.034 | 0.026 | 0.017 |
|  | SSIM ↑ | 0.864 | 0.897 | 0.929 |
|  | PSNR ↑ | 23.61 | 25.40 | 27.19 |
| Random Masks | MAE ↓ | 0.013 | 0.015 | 0.017 |
|  | SSIM ↑ | 0.898 | 0.910 | 0.925 |
|  | PSNR ↑ | 25.68 | 27.48 | 29.27 |
| Object Removal | MAE ↓ | 0.064 | 0.035 | 0.019 |
|  | SSIM ↑ | 0.842 | 0.869 | 0.896 |
|  | PSNR ↑ | 21.78 | 23.52 | 25.26 |

## 4 COMPARISONS & DISCUSSION

In this chapter, both qualitative & quantitative comparative analyses of our research work are demonstrated. The comparative analysis is between some famous works and our work.

### 4.1 Qualitative Comparison

We have compared our method to the state-of-the-art for guided picture inpainting and found that it is more accurate and efficient. Figure 11 shows that CA [16] produces significant artifacts that mislead the structure. While SF [44] reconstructs the missing region with proper textures since it preserves both structures and textures, though it has several noticeable inconsistencies towards the edges.GC [26] reconstructs more realistic pictures since this method contains convolution layers and also the refinement network, though it misses out on notable texture features. TM [25] generates better results than the previous method, though it smoothens the masked area; as a result, the images lose their texture. Compared to other approaches, Our method generates more realistic and accurate textures in the corrupted regions.

For more comparison, we have tested our model on the places2 dataset; here, we have also compared our approach with previous well-known methods. As seen from Fig. 12, Global Local [19] generates artifacts that misrepresented the picture, CA [16] produces fewer artifacts, though still visually misrepresented the architecture, while PC [17] produces better results, though still appears noticeable color discrepancy. GC [26] generates good results here, though it smoothens the mask area, where our method gives more accurate and realistic inpainting than others.

For blind image inpainting, we have compared the previously well-known model on Places and CelebA datasets. TM. As seen in Fig. 14, TM [25] does not produce clouds in the sky, though ground truth has clouds in the sky in the first row. From the second row of Fig. 14, TM [25] couldn't generate any curve of the tower, whereas our method generates the curve of the tower, which is closer to ground truth. For blind face inpainting, our model generates more realistic faces than the previous state-of-the-art methods.

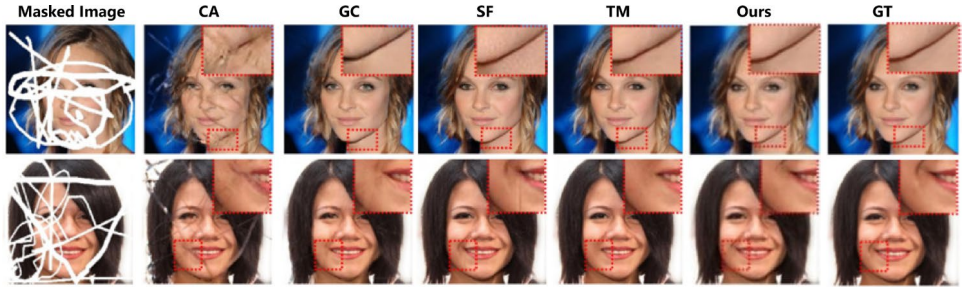

**Fig. 11**: Comparison of guided inpainting on CelebA dataset from left to right masked input, CA [16], GC [26], SF [44], TM [25], our result and ground truth images

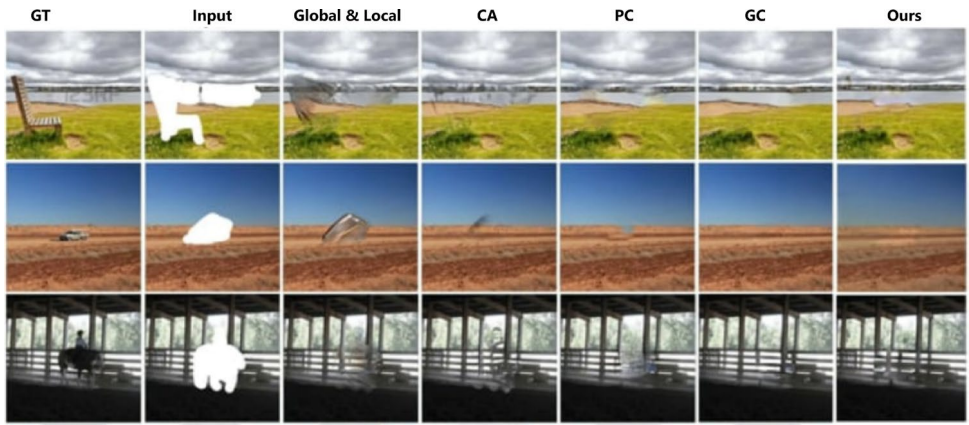

**Fig. 12**: Darawing a qualitative comparison between our model and the state-of-the-art models, by utilizing a random mask to hide the objects to be deleted: from left to right Ground Truth image, input, Global & Local [19], CA [16], PC [17], GC [26] and our result.

We've qualitatively compared our model to the Places2 dataset for object removal. From Fig. 13, Yu et al. [16] generate visual artefacts on the removed object. GC [26] performed better but still appears to have colour incongruity in the masked region, while our model is more realistic, as there is no object in the masked region.

### 4.2 Quantitative Comparison

We've quantitatively compared our approach using three well-known evaluation metrics and three datasets. The detailed specification of datasets is already mentioned in the Dataset Exploration subsection of the Methodology Section. We use the same masks to achieve a fair comparison.

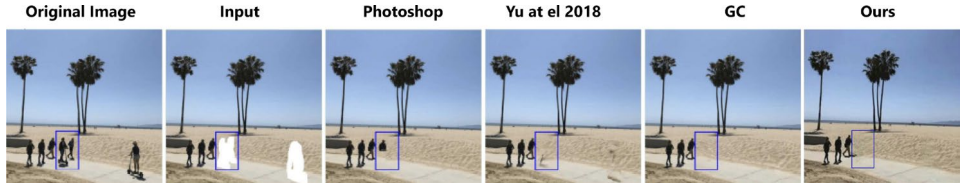

**Fig. 13**: An example of qualitative comparison of object removal between our method and state-of-the-art approaches: From left Ground Truth image, input, Adobe Photoshop, Yu [16], GC [26], and our result.

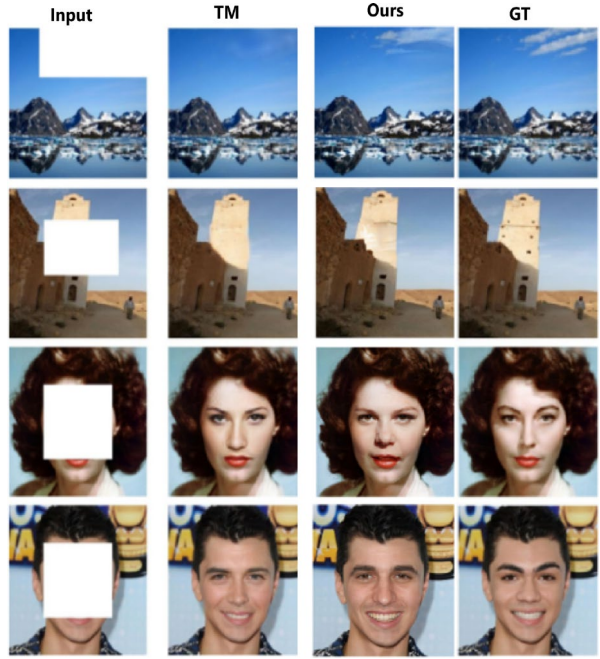

**Fig. 14**: Qualitative comparison using a rectangular mask blind inpainting on Places2 and CelebA datasets: From left to right input, TM [25], our result, and lastly Ground Truth (GT).

#### 4.2.1 Free-Mask Inpainting with Randomized H2I Ratios

In image inpainting, one critical challenge is handling missing regions that vary in size and shape. Free-mask inpainting, where the hole-to-image (H2I) ratios are randomized, tests the adaptability and robustness of models in these complex scenarios. Unlike fixed masks, which are predictable and uniform, free masks create a more realistic situation, mimicking real-world images where missing areas can appear anywhere and in any form.

Table 6 presents the quantitative results for the Places2 dataset. We can see that CA [16] performs the weakest across all three metrics. EC [24] shows some improvement, likely due to its edge-prediction capabilities, which help in creating more defined image structures. DFNet [51] and GC [26] have very similar scores, indicating comparable performance. SF [44] and TM [25] outperform these earlier methods, with results that are closely aligned with each other. However, our proposed model clearly stands out as the best among all the approaches, delivering superior results in all metrics.

Moving on to the CelebA dataset, as shown in Table 7, CA [16] again performs relatively less than other methods. EC [24] performs better than CA [16] but still falls short compared to more advanced models. DFNet [51], GC [26], SF [44], and TM [25] deliver closely matched results, performing reasonably well. Once again, our approach significantly outperforms these methods, demonstrating its effectiveness across different scenarios.

**Table 6**: Performance evaluation with significant parameters in the Places2 [47] dataset

| Places2 | CA [16] | EC [24] | DFNet [51] | GC [26] | SF [44] | TM [25] | Ours(Proposed) |
|---|---|---|---|---|---|---|---|
| MAE ↓ | 0.075 | 0.053 | 0.045 | 0.045 | 0.044 | 0.042 | **0.036** |
| SSIM ↑ | 0.660 | 0.762 | 0.803 | 0.810 | 0.812 | 0.816 | **0.902** |
| PSNR ↑ | 18.27 | 20.80 | 21.03 | 21.78 | 21.97 | 22.55 | **23.19** |

**Table 7**: Performance evaluation with significant parameters on CelebA [46] dataset

| CelebA | CA [16] | EC [24] | DFNet [51] | GC [26] | SF [44] | TM [25] | Ours(Proposed) |
|---|---|---|---|---|---|---|---|
| MAE ↓ | 0.086 | 0.066 | 0.052 | 0.052 | 0.049 | 0.045 | **0.041** |
| SSIM ↑ | 0.560 | 0.662 | 0.703 | 0.710 | 0.712 | 0.716 | **0.815** |
| PSNR ↑ | 21.03 | 23.17 | 24.30 | 25.06 | 25.633 | 26.64 | **27.19** |

### 4.2.2 Free-Mask Inpainting with Different H2I Ratios

Evaluating the performance of image inpainting models using different hole-to-image (H2I) area ratios provides valuable insight into their adaptability across various mask sizes. By analyzing these results, we aim to understand how well each method handles increasingly complex scenarios where larger portions of the image need to be reconstructed.

**Analysis of the Places2 Dataset**

Table 8 shows the quantitative comparison of various inpainting techniques on the Places2 dataset. For the smallest H2I ratio (0.01, 0.1], the proposed method achieves the highest scores across all three metrics, surpassing the existing approaches significantly. It reaches a PSNR of 35.87 and an SSIM of 0.9897, clearly outperforming its closest competitor, CMT [38]. As the H2I ratio increases to (0.1, 0.2], our approach continues to lead with a PSNR of 29.76 and an SSIM of 0.9571.

**Table 8**: Quantitative comparison on the Places2 [47] dataset according to the hole-to-image (H2I) area ratios.

| Method | H2I ∈ (0.01, 0.1] | | | H2I ∈ (0.1, 0.2] | | |
|---|---|---|---|---|---|---|
| | PSNR ↑ | SSIM ↑ | FID ↓ | PSNR ↑ | SSIM ↑ | FID ↓ |
| EdgeConnect [24] | 33.80 | 0.9811 | 3.41 | 28.41 | 0.9524 | 7.98 |
| ICT [36] | 31.66 | 0.9771 | 3.98 | 28.45 | 0.9446 | 9.96 |
| BAT [37] | 34.54 | 0.9839 | 2.49 | 29.01 | 0.9499 | 6.84 |
| MAT [35] | 34.43 | 0.9838 | 2.41 | 28.08 | 0.9549 | 6.19 |
| CMT [38] | 35.43 | 0.9845 | 2.29 | 29.28 | 0.9546 | 5.93 |
| Proposed | **35.87** | **0.9897** | **2.17** | **29.76** | **0.9571** | **5.04** |

| Method | H2I ∈ (0.2, 0.3] | | | H2I ∈ (0.3, 0.4] | | |
|---|---|---|---|---|---|---|
| | PSNR ↑ | SSIM ↑ | FID ↓ | PSNR ↑ | SSIM ↑ | FID ↓ |
| EdgeConnect [24] | 25.29 | 0.9148 | 13.56 | 23.12 | 0.8744 | 18.90 |
| ICT [36] | 24.43 | 0.9094 | 15.82 | 21.01 | 0.8547 | 22.90 |
| BAT [37] | 26.43 | 0.9263 | 11.26 | 21.85 | 0.8688 | 17.39 |
| MAT [35] | 24.62 | 0.9155 | 10.79 | 22.00 | 0.8721 | 20.16 |
| CMT [38] | 25.88 | 0.9240 | 10.36 | 23.56 | 0.8850 | 14.69 |
| Proposed | **26.39** | **0.9286** | **9.75** | **22.68** | **0.8904** | **12.73** |

We see a similar trend when analyzing the H2I ratios in the range (0.2, 0.3] and (0.3, 0.4]. EdgeConnect [24] and ICT [36] struggle with a noticeable drop in both PSNR and SSIM values, while our method remains robust with a PSNR of 26.39 and 22.68 respectively, and consistently achieves the best FID scores. The proposed method's resilience across these ratios highlights its ability to maintain image quality even with larger missing areas.

**Analysis of the CelebA Dataset**

As shown in Table 9, the CelebA dataset results also demonstrate the superior performance of our proposed model. For the H2I range of (0.01, 0.2], the PSNR score for our approach reaches 36.20, with an SSIM of 0.9864, significantly higher than those of other models like MAT [35] and CMT [38]. Even when the ratio increases to (0.2, 0.4], our method continues to excel, achieving a PSNR of 29.18 and an FID of 2.49, which clearly indicates fewer perceptual discrepancies compared to other approaches.

The noticeable gap between our model and the other techniques suggests that our proposed method is effective in handling minor defects and robust in larger-scale inpainting tasks. This consistency across different H2I ratios points to its enhanced capability to generate high-quality reconstructions, regardless of the complexity of the missing regions.

**Analysis of the ImageNet Dataset**

Moving on to the ImageNet dataset results shown in Table 10, the proposed method again demonstrates its strength. For the H2I range of (0.2, 0.4], it leads with a PSNR of 24.918 and an SSIM of 0.921, outperforming both PIC [31] and ICT [36]. This trend continues even with the more challenging (0.4, 0.6] ratio, where our approach achieves the highest PSNR of 29.64 and the best FID score of 45.38.

**Table 9**: Quantitative comparison on the CelebA [46] dataset according to the hole-to-image (H2I) area ratios.

| Method | H2I ∈ (0.01, 0.2] | | | H2I ∈ (0.2, 0.4] | | |
|---|---|---|---|---|---|---|
| | PSNR ↑ | SSIM ↑ | FID ↓ | PSNR ↑ | SSIM ↑ | FID ↓ |
| ICT [36] | 33.27 | 0.9793 | 1.87 | 26.40 | 0.9389 | 5.61 |
| BAT [37] | 34.63 | 0.9830 | 1.06 | 26.91 | 0.9440 | 3.75 |
| MAT [35] | 35.31 | 0.9842 | 0.90 | 27.67 | 0.9461 | 2.55 |
| CMT [38] | 35.92 | 0.9859 | 0.84 | 28.24 | 0.9515 | 2.54 |
| Proposed | **36.20** | **0.9864** | **0.81** | **29.18** | **0.9533** | **2.49** |

**Table 10**: Quantitative comparison on ImageNet [48] dataset according to the hole-to-image (H2I) area ratios.

| Method | H2I ∈ (0.2, 0.4] | | | H2I ∈ (0.4, 0.6] | | |
|---|---|---|---|---|---|---|
| | PSNR ↑ | SSIM ↑ | FID ↓ | PSNR ↑ | SSIM ↑ | FID ↓ |
| PIC [31] | 24.010 | 0.867 | 47.750 | 18.843 | 0.642 | 101.278 |
| ICT [36] | 24.757 | 0.888 | 28.818 | 20.135 | 0.721 | 59.486 |
| Proposed | **24.918** | **0.921** | **17.49** | **29.64** | **0.811** | **45.38** |

Overall, the analysis across the Places2, CelebA, and ImageNet datasets demonstrates the consistent superiority of our proposed method in free-mask inpainting with varying H2I ratios. The results indicate that our approach handles small to medium missing regions effectively and scales well when dealing with larger holes in the images. This adaptability and robustness make it a compelling choice for practical applications in image restoration tasks.

### 4.2.3 Fixed-Mask Inpainting

We have used fixed masks to test our model's performance with consistent and predefined missing regions in images.

As shown in Table 11, our proposed method demonstrates the best results across all metrics on the Places2 dataset with fixed masks. It achieves the highest PSNR of 22.66, indicating superior detail reconstruction compared to CA [16], EC [24], and TM [25]. Our model also leads in SSIM with a score of 0.839, showcasing its strength in maintaining the structural integrity of the inpainted images.

Additionally, with the lowest FID score of 7.70, our approach outperforms other models in generating images with minimal artifacts, resulting in more realistic and natural visuals.

### 4.3 Object Removal

The user can remove any unwanted objects from the image by interactively drawing the mask. Our proposed hybrid framework can recover the corrupted parts without any artifact to maintain the proper texture of the image. In Fig. 13, we've qualitatively compared our model with the previously well-known approach in Places2. The comparison shows that Photoshop is unable to remove the entire object. Yu et al. [16] has a visual artifact in the corrupted region. The GC [26] showed a better result but still has a color deficiency in the reconstructed image. However, our approach has shown a very competitive result in recovering all the corrupted parts by maintaining proper contrast and textures.

**Table 11**: Quantitative comparison on the Places2 [47] dataset for fixed-mask

|        | CA [16] | EC [24] | TM [25] | Ours(Proposed) |
|--------|---------|---------|---------|----------------|
| PSNR ↑ | 20.65   | 21.75   | 21.98   | **22.66**      |
| SSIM ↑ | 0.818   | 0.823   | 0.835   | **0.839**      |
| FID ↓  | 8.31    | 8.16    | 7.99    | **7.70**       |

## 4.4 Performance Parameter Calculation

Since our dataset contains more than 0.5 million images, only the accuracy parameter is not enough to measure our model's performance. So, we have calculated recall, precision, and F1-score.

**Recall:** It means that out of the total positive actual values, how many positive values were detected correctly. We calculated the recall or true positive rate per equation 6.

$$Recall = \frac{TruePositive}{TruePositive + FalseNegative} \qquad (6)$$

**Precision:** It means how many were actual positive values out of total positive predicted values. In addition, we have calculated the precision or positive predictive value as stated in Equation 7.

$$Precision = \frac{TruePositive}{TruePositive + FalsePositive} \qquad (7)$$

**PSNR** is used to compute the noise ratio between two images. This parameter is used to measure the quality between the original image and the reconstructed image. The higher the value of PSNR, the better the quality of the inpainted image. PSNR is expressed by equation 8.

$$PSNR = 10 \log_{10} \frac{PeakValue^2}{MSE} \qquad (8)$$

**F1-score:** F1-score is the harmonic mean of recall and precision. It evaluates both false positives and false negatives. For uneven classification, it is more useful than accuracy. It is stated by equation 9

$$F1 - Score = 2 * \frac{Precision * Recall}{Precision + Recall} \qquad (9)$$

The **SSIM** is a perception-based model. In this way, image degeneration is understood as a shift in the perception of structural information. Pixels that are considerably interconnected or spatially limited are referred to as structural information. These highly interlinked pixels hint at more critical information about visual items in the image realm. Luminance masking is the technique of making a picture's distortion component less evident around the image's edges. Contrast masking, on the other hand, refers to the practice of making textural distortions in a picture less visible. SSIM calculates the perceived quality of photos and videos. It assesses the similarity between the ground truth and the reconstructed image.

## 4.5 Limitations of the study

While the proposed hybrid multiGAN-based image inpainting architecture shows promising results, there are a few fundamental limitations to consider. Firstly, the model's computational complexity and extended training time due to the integration of multiple GAN networks make it less suitable for real-time applications or scenarios with limited computational resources. Additionally, while the model performs well on datasets like CelebA, Places2, and ImageNet, its ability to generalize to entirely new or unseen datasets remains uncertain and may require further fine-tuning. These limitations highlight areas for future improvement, particularly in enhancing the model's efficiency and generalization capabilities.

# 5 CONCLUSION

In this study, we present a novel hybrid multiGAN-based extreme image inpainting architecture, which encompasses two fundamental networks: one for texture and structure regeneration and the other for color and context regeneration. The first model, referred to as Transpose convolution-based GAN is responsible for coarse regeneration, employing advanced generators for stable training and an intelligent discriminator to distinguish between real and locally generated outputs. Co-Mod GAN serves as a refinement network, taking transpose convolution-based GAN's output and producing global refined output. Additionally, FR-CNN is employed to remove and add objects locally to an image, and its output is also fed to Co-Mod GAN for global refinement. This approach demonstrates promising results for image inpainting in both user-guided and blind inpainting scenarios involving large missing areas. During the training phase, three types of losses are applied to the three networks, with each model separately trained using the CelebA, Places2, and ImageNet datasets. Major findings of the proposed research include:

- Comprehensive coverage of various aspects of image inpainting (guided, blind image inpainting, and object removal with inpainting).
- Capability to remove objects from images using a user-guided mask.

Future research will explore image out-painting using the proposed method to achieve these results without human intervention. We also plan to investigate image forensics of inpainted images and consider video inpainting applications.

## Statements and Declarations

**Conflict of interest.** The authors declare that they have no conflict of interest.

**Data Availability Statement.** Data supporting Fig. 8, 9, 10, 11, 12, 13, 14 and Tables 2, 3, 4, 5, 6, 7, 8, 9, 10, 11 are publicly available datasets: CelebA dataset [46] is available at https://doi.org/10.1109/ICCV.2015.425, Places2 dataset [47] is available at https://doi.org/10.1109/TPAMI.2017.2723009 & Image-Net dataset [48] is available at https://doi.org/10.1109/CVPR.2009.5206848.